\pdfoutput=1
\documentclass{article}

\usepackage{microtype}
\usepackage{graphicx}
\usepackage{subcaption}
\usepackage{booktabs}
\usepackage{hyperref}

\usepackage[preprint]{icml2026}
\hypersetup{pdfsubject={}}

\usepackage{amsmath,amsfonts,bm}

\def\eqref#1{equation~\ref{#1}}

\def\1{\bm{1}}

\DeclareMathAlphabet{\mathsfit}{\encodingdefault}{\sfdefault}{m}{sl}
\SetMathAlphabet{\mathsfit}{bold}{\encodingdefault}{\sfdefault}{bx}{n}

\usepackage{url}
\usepackage{amsmath}
\usepackage{amssymb}
\usepackage{mathtools}
\usepackage{amsthm}
\usepackage{epsfig}
\usepackage{multirow}
\usepackage{xspace}
\usepackage{fancyhdr}
\usepackage{color}
\usepackage{colortbl}
\usepackage{xcolor}
\definecolor{mygray}{gray}{.85}
\definecolor{myyellow}{RGB}{204,102,0}
\definecolor{myred}{RGB}{204,0,102}
\definecolor{mypink}{RGB}{234,182,203}
\definecolor{mypurple}{RGB}{226,227,239}
\definecolor{maroon}{cmyk}{0,0.87,0.68,0.32}
\definecolor{myblue}{RGB}{189,223,254}
\definecolor{mygrey}{RGB}{228,228,228}

\usepackage{mathrsfs}
\usepackage{arydshln}
\usepackage{siunitx}
\usepackage{wrapfig}
\usepackage{appendix}
\usepackage{enumitem}
\usepackage{float}
\usepackage{placeins}

\newcommand{\model}{InstructPro2\xspace}
\newcommand{\modelAB}{InstructPro2-Ab\xspace}
\newcommand{\modelLIG}{InstructPro2-Lb\xspace}
\newcommand{\dataset}{InstructPro2Bench\xspace}

\icmltitlerunning{InstructPro2: Natural Language and Biological Grounding Guided De Novo Protein Design}

\begin{document}

\twocolumn[
\icmltitle{InstructPro2: Natural Language and Biological Grounding\\
           Guided \textit{De Novo} Protein Design}

\icmlsetsymbol{equal}{*}

\begin{icmlauthorlist}
\icmlauthor{Charles Novak}{equal,cbd}
\icmlauthor{Zhenqiao Song}{equal,lti}
\icmlauthor{Chuan Li}{lambda}
\icmlauthor{Jianwen Xie}{lambda}
\icmlauthor{Lei Li}{lti}
\end{icmlauthorlist}

\icmlaffiliation{cbd}{Ray and Stephanie Lane Computational Biology Department, School of Computer Science, Carnegie Mellon University, Pittsburgh, PA 15213, USA}
\icmlaffiliation{lti}{Language Technologies Institute, School of Computer Science, Carnegie Mellon University, Pittsburgh, PA 15213, USA}
\icmlaffiliation{lambda}{Lambda Inc, San Francisco, CA 94105, USA}

\icmlcorrespondingauthor{Lei Li}{leili@cs.cmu.edu}

\icmlkeywords{Machine Learning, Protein Design}

\vskip 0.3in
]

\printAffiliationsAndNotice{\icmlEqualContribution}

\begin{abstract}
The ability to generate proteins with desired functions is a central goal of
computational biology. Although natural language can express diverse design
objectives, existing protein generation methods often support only narrow
tasks or make limited use of biological constraints. We develop \model to
translate natural-language design objectives into protein sequences while
incorporating task-specific biological context. A text encoder captures the
design instructions, while dedicated ligand and antigen encoders provide
biological context to the sequence-generating decoder. For pretraining, we
assemble \dataset, comprising 205.7 million $\langle$function description,
protein$\rangle$ pairs from UniProtKB. We subsequently adapt the model to
ligand-binding protein and antibody design. On functional protein generation
benchmarks, \model outperforms Pinal, the strongest baseline, achieving a
functional recovery F1 of 74.8 versus 62.2 and lowering global RMSD from
5.33\AA\ to 3.04\AA. For antibody design, the model obtains a DockQ score of
0.39, while ligand-binding protein design reaches a docking success rate of
72.6\%. These findings highlight how integrating language-based instructions
with biological conditioning can extend protein generation to distinct
design settings.

\end{abstract}

\section{Introduction}
Proteins execute a vast range of functions essential to biological life. The ability to design functional proteins with specified properties on demand holds transformative potential across numerous domains, including therapeutic development \citep{nelson2010development}, enzyme engineering for sustainable chemical production \citep{bornscheuer2012engineering}, and the creation of novel diagnostic tools and biosensors \citep{rottinghaus2022engineering}. By enabling direct, targeted generation, efficient computational protein design circumvents the heavy reliance on slow natural evolutionary processes and labor-intensive high-throughput screening.

Recent progress has brought the field significantly closer to this goal. Large protein language models trained on evolutionary sequence data have captured much of the underlying grammar of proteins, enabling the generation of plausible, well-folded sequences~\citep{nijkamp2022progen2,ferruz2022protgpt2,lin2023esm2}. Building on these foundations, generative frameworks have introduced explicit conditioning into protein design via specified structural motifs~\citep{watson2023novo} or functional tags~\citep{madani2023large}. Most recently, several systems have extended this paradigm to support natural language prompts for desired functions, making protein design increasingly versatile~\citep{dai2024pinal,song2025instructpro}. However, despite their effectiveness, existing methods remain constrained by narrow domain specificity, restricted design scopes, or insufficient biological grounding, which ultimately limits their design success rates.

To address these challenges, we introduce \model, a functional protein design framework that integrates natural language instructions with contextual biological knowledge. The architecture comprises a text encoder for natural language prompts, dedicated biological encoders (specifically for ligands and antigens) to capture biological grounding context, and a protein decoder that, conditioned on both modalities, generates protein sequences aligned with the specified function and binding target. We employ a two-stage training strategy: first, we pretrain the model on instruction-following tasks to establish broad generalization across design objectives; second, we incorporate biological context to optimize performance on real-world downstream tasks.

Concretely, we make the following contributions:
\begin{itemize}[nosep,leftmargin=2.0em]
\item We introduce a text-conditioned protein design model capable of generating diverse, functional proteins directly guided by human natural language instructions. Our code is available here: \url{https://github.com/chovak/InstructPro2}.
\item We curate \dataset, a large-scale dataset comprising 205.7 million $<$function description, protein$>$ pairs across diverse families to support model pretraining.
\item We pretrain \model on \dataset, demonstrating that it outperforms
strong baselines, including ESM3~\citep{hayes2025esm3},
ProGen2~\citep{nijkamp2022progen2}, and Pinal~\citep{dai2024pinal},
achieving a lower global RMSD of 3.04\,\AA\ (vs.\ 5.33\,\AA\ for the
strongest baseline) and a higher functional recovery F1 of 74.8 (vs.\
62.2).
\item We fine-tune the pretrained model on two key downstream tasks, showing that the text-conditioned decoder seamlessly transfers to antibody design (achieving a DockQ score of 0.39) and ligand-binding protein design (achieving a 72.6\% docking success rate).
\end{itemize}

\section{Related Work}
\textbf{Functional Protein Design.} Designing proteins with desired functions has been studied with a wide range of methods. Early efforts demonstrated that models trained on large-scale evolutionary sequences can generate diverse, well-folded proteins without explicit conditioning \citep{nijkamp2022progen2, ferruz2022protgpt2}. Subsequent studies introduced conditioning signals to steer generation toward specific attributes, structures, or binding targets \citep{krishna2024rosettafoldaa, zambaldi2024novo}. Specifically, tag-based approaches leverage functional annotations, such as protein families \citep{madani2023large}, Enzyme Classification (EC) numbers \citep{munsamy2022zymctrl}, or species and chain types \citep{shuai2023iglm} to guide generation toward targeted protein classes. Parallel work focuses on structure-conditioned generation, designing proteins that fold into specified backbones \citep{dauparas2022robust, hsu2022learning} or scaffold key functional motifs \citep{watson2023novo, wang2022scaffolding}. Most recently, advanced models enable atomic-level design tailored to designated binding partners \citep{stark2025boltzgen, butcher2025rfdiffusion3, didi2026scaling}. Despite these advances, the efficient and precise design of proteins with arbitrary, user-defined functions remains an open challenge.

\textbf{Natural Language-Guided Protein Design.} Inspired by the instruction-following capabilities of large language models~\citep{ouyang2022training,wang2023self,zhou2023instruction}, recent research has begun leveraging natural language conditioning for protein design. One prominent approach maps textual descriptions to intermediate structural representations, which are subsequently translated into protein sequences via generative models \citep{dai2024pinal, praljak2024biom3, riley2025mp4}. Other methods learn joint protein--text embedding spaces, enabling sequence retrieval or generation based on text representations \citep{wu2024proteinclip, wang2024instructprotein, liu2025text}. More recent efforts train models to design functional proteins directly from open-ended human instructions \citep{song2025instructpro}. However, existing methods remain restricted to narrow design tasks and limited real-world scenarios. Developing a generalizable, instruction-guided model capable of generating biologically grounded proteins for arbitrary, user-defined goals remains an open challenge.

\section{Proposed Method: \model}
\begin{figure*}[t]
\centering
\includegraphics[width=\textwidth]{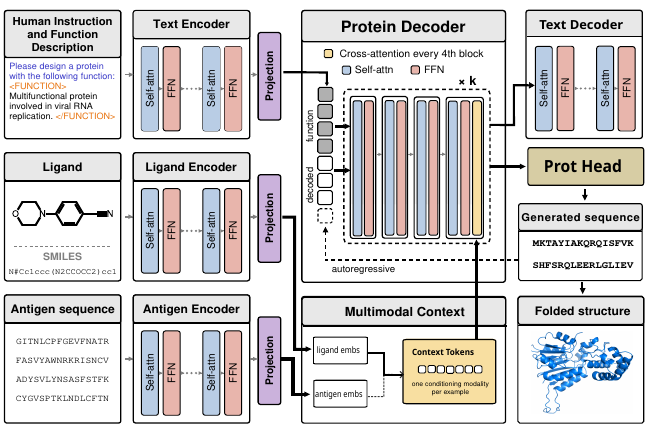}
\caption{Overview of the \model architecture. \textbf{(a) During pretraining}: natural-language function descriptions are encoded by the text encoder, and then the protein decoder takes the encoded representations as prefix to generate a consistent protein sequence. 
\textbf{(b) During fine-tuning:} biological grounding is incorporated alongside textual representations to enable contextually realistic protein design for real-world applications, such as ligand binding and antibody engineering.
}
\label{fig:architecture}
\end{figure*}

\subsection{Problem Definition}
Let $r \in \mathcal{A}^{N_r}$ denote a protein sequence over an amino-acid vocabulary $\mathcal{A}$, and $w \in \mathcal{V}^{N_w}$ represent a natural-language functional description over a text vocabulary $\mathcal{V}$, where $N_r$ and $N_w$ denote their respective sequence lengths. We represent the biological grounding as $C$, which captures context-specific constraints such as a target ligand's SMILES string or an antigen sequence. Our objective is to learn a generative model $p_\theta(r \mid w, C)$, parameterized by $\theta$, to design functional protein sequences $r$ that follow natural-language instructions $w$ while satisfying the biological grounding $C$.

\subsection{Model Architecture}
\label{sec:architecture}
The overall architecture of \model is illustrated in Figure~\ref{fig:architecture}. \model integrates a text encoder to process natural-language functional descriptions alongside specialized biological encoders, specifically a ligand encoder and an antigen encoder to capture contextual biological grounding. Conditioned on these representations, the protein decoder generates a protein sequence that satisfies the specified functional requirements while effectively binding to the targeted biological partner.

\textbf{Protein decoder and text conditioning.} At the core of \model is a protein decoder, a GPT-J-style autoregressive transformer with rotary position embeddings. We
encode the natural-language instructions $w$ with an autoregressive text encoder,
\begin{equation}
H_w = \mathrm{Enc}_{\text{text}}(w), \qquad \tilde H_w = W_p H_w,
\end{equation}
and concatenate the projected representation $\tilde H_w$ as a prefix to
the protein token embeddings $E(r)$. A single block-causal pass through
the decoder then yields the protein sequence logits,
\begin{equation}
p_\theta(r_i \mid r_{<i}, w, C) = \mathrm{softmax}\big(W_o X_L^{(i)}\big),
\end{equation}
where $X_L^{(i)}$ is the decoder's final hidden state at target position
$i$. A second, text decoder takes the
decoder's hidden states at the text positions, $X_L^{w}$, and reconstructs
the conditioning description from them in a single teacher-forced pass,
giving an auxiliary signal that keeps the decoder from discarding the
description after the first pass.

\textbf{Modality conditioning.} For the antibody and ligand variants, frozen modality encoders $\phi$ encode the corresponding auxiliary modality $c$. The resulting representation is projected into the decoder's
embedding space by $W_c$, rescaled by a learned per-modality scalar
$s_m \in \mathbb{R}$, and offset by a learned modality-type embedding $e_m$:
\begin{equation}
\label{eq:eq3}
C = s_m \cdot W_c\,\phi(c) + e_m.
\end{equation}
We insert zero-initialized gated cross-attention adapters \citep{alayrac2022flamingo} every four
decoder layers, for eight adapters in total. Writing $R_\ell$ for the output of decoder block
$\ell$'s own sublayers,
\begin{equation}
R_\ell = X_\ell + \text{SelfAttn}(\text{LN}(X_\ell)) + \text{MLP}(\text{LN}(X_\ell)),
\end{equation}
the adapter appended to that block computes
\begin{equation}
X_{\ell+1} = R_\ell + \tanh(\alpha_\ell)\cdot\text{CrossAttn}(\text{LN}(R_\ell), C, C).
\end{equation}
Here $\text{LN}$ denotes layer normalization, $X_\ell$ the hidden states
entering decoder block $\ell$, and $R_\ell$ the output of that block's
self-attention and feed-forward sublayers. The cross-attention takes
$\text{LN}(R_\ell)$ as queries and the multimodal context $C$ as keys and
values. Since $\alpha_\ell$ is initialized to zero, this modality
pathway initially contributes no signal, leaving the pretrained decoder's
computation unchanged at the start of finetuning. During pretraining, all components are trained jointly: the protein decoder, the encoder layers, the text projection, and the reconstruction head. During fine-tuning, we initialize from the pretraining checkpoint and add the modality projections and gated cross-attention adapters. All of these remain trainable during finetuning, while the text encoder and the two modality encoders are held frozen, so that the conditioning representations stay fixed and the model learns to read them rather than reshape them.

\textbf{Training objective.} Across both stages, \model minimizes
\begin{equation}
\begin{aligned}
\mathcal{L}_{\text{protein}} &= -\sum_{i=1}^{N_r} \log p_\theta(r_i \mid r_{<i}, w, C), \\
\mathcal{L}_{\text{text}} &= -\sum_{j=1}^{N_w} \log p_\theta(w_j \mid w_{<j}),
\end{aligned}
\end{equation}
\begin{equation}
\mathcal{L} = \mathcal{L}_{\text{protein}} + \lambda\,\mathcal{L}_{\text{text}}.
\end{equation}
$C$ is omitted during the pretraining stage. The text reconstruction term is
weighted by a hyperparameter $\lambda$, which keeps it an auxiliary signal so
that protein generation stays the primary objective. We set $\lambda = 0.5$
in our experiments.

\subsection{\dataset Construction}
\label{sec:dataset}
To train our foundation model, we constructed \dataset, a large-scale corpus of
205,727,986 protein-text pairs. A central challenge in constructing such a
corpus is obtaining functional descriptions at the scale of hundreds of
millions of proteins without sacrificing the biological specificity of the
annotations. We address this by grounding natural-language descriptions in
InterPro annotations \citep{blum2025interpro}, which provide a structured
vocabulary of protein domains, sites, motifs, families, and other sequence
features that are directly associated with protein function. We summarize the 49,677 InterPro entries once using Gemini 3.1 Pro \citep{googledeepmind2026gemini31pro}, producing reusable natural-language components that can be composed across proteins by combining the functional features associated with each protein.

To assemble the functional description for each protein, we select up to three
InterPro entries associated with the protein whose descriptions together
capture distinct functional characteristics. We prioritize more specific functional annotations over broader ones, selecting
sites, active sites, and PTMs first, followed by domains and repeats, families,
and superfamilies. We then remove redundant annotations among the selected entries using Jaccard
similarity, allowing the remaining selections to capture distinct aspects of the
protein's function. The selected descriptions are then composed into a single natural-language instruction prompt using templates that provide connector phrases between individual functional components, producing a consistent, composable representation of protein function for training.

We then train on the resulting dataset using two complementary objectives, enabling the model to both generate complete protein sequences from function and complete partially specified sequences conditioned on function:
\begin{itemize}
    \item \textbf{Autoregressive generation} ($\sim$50\% of examples): The
    model receives the natural-language function description and is tasked
    with generating the full protein sequence autoregressively.
    \item \textbf{Masked infilling} ($\sim$50\% of examples): We apply a span-corruption objective, akin to T5 \citep{raffel2020t5} and SpanBERT \citep{joshi2020spanbert}, in which 20--40$\%$ of each protein sequence is masked across multiple spans. The model reconstructs the masked regions conditioned on the surrounding sequence and function description. This same span-infilling formulation is reused for HCDR generation in the antibody design task, providing a shared mechanism for localized sequence design across training stages. Further details of the masking and span construction procedure are provided in Appendix~\ref{app:masking}.
\end{itemize}

\textbf{Clustering and splits.} To prevent data leakage between training
and evaluation, we divide the dataset strictly at the cluster level. We
cluster the sequences using MMseqs2 \citep{steinegger2017mmseqs2} at 50$\%$
sequence identity and 80$\%$ coverage, producing 29,008,229 distinct clusters.
We assign approximately 1,000 clusters each to the validation and test sets.
The training split uses all proteins within its assigned clusters, while
validation and test evaluate a single representative sequence per cluster,
selected via reservoir sampling.

\begin{figure*}[t!]
\centering
\includegraphics[width=\textwidth]{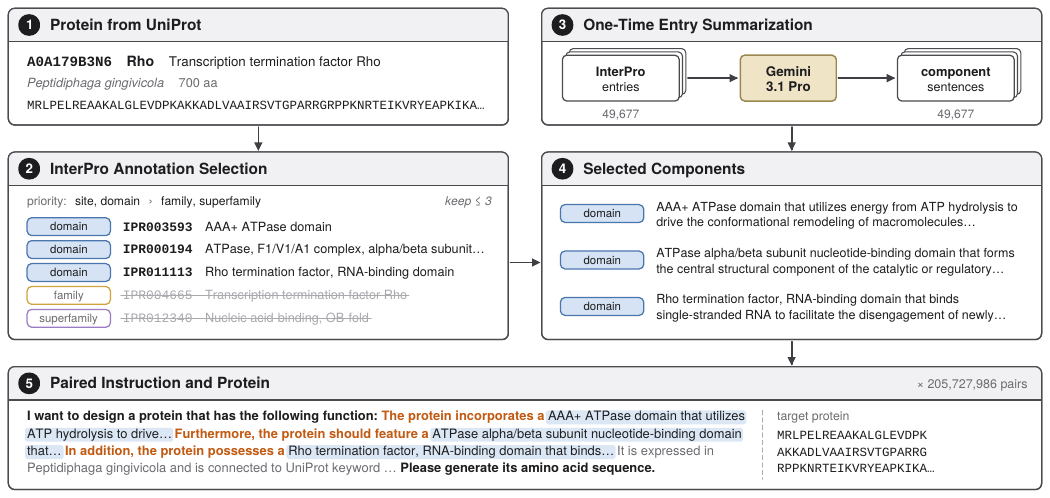}
\caption{Automated pipeline to construct \dataset
}

\label{fig:interpro_composability}
\end{figure*}

\subsection{Real-World Task Data Curation}
\label{sec:stage2}
\textbf{Antibody-antigen design blueprints.}
To extend natural-language-guided protein design to antibodies, we construct examples combining antibody sequences, antigen information, and natural-language descriptions. We begin with a high-quality subset of experimentally solved SAbDab \citep{dunbar2014sabdab} antibody-antigen complexes for which both antibody and antigen structures are available. Using the structural and functional information associated with each complex, we generate natural-language descriptions of the target antigen and intended antibody function. Because the paired antibody-antigen dataset is limited in scale, we supplement it with additional antibody examples without paired antigen information from SAbDab, TrEMBL, and SwissProt \citep{uniprot2025}. For these examples, available antibody annotations are converted into natural-language descriptions using templates, providing additional sequence-function training data without requiring antigen conditioning.

Because CDR-H3 is the primary design target, we cluster the antibody
examples by CDR-H3 sequence identity to prevent sequence-level leakage
between training and evaluation, and perform the train,
validation, and test split at the cluster level. This ensures that closely
related CDR-H3 sequences are not distributed across different splits.

\textbf{Ligand-protein design dataset.} To extend natural-language-guided
protein design to ligand-binding proteins, we construct protein-ligand-text
triples pairing a target ligand and a natural-language function description
with a protein sequence with known binding to the ligand. We build this
dataset from the train, validation, and test splits of InstructProBench
\citep{song2025instructpro}, giving 8,727,514 training examples, 3,263
validation examples, 1,500 seen-ligand test examples, and 127 unseen-ligand
test examples held out to evaluate generalization to ligands not observed
during training. Given the scale of this dataset, we construct each
protein's function description using the same InterPro-grounded pipeline
used for pretraining. The ligands are represented as SMILES strings and encoded
with a frozen, pretrained ChemBERTa encoder \citep{chithrananda2020chemberta},
whose representations are provided to the model through the cross-attention
adapters described in Section~\ref{sec:architecture}.

\section{Experiments}
\label{sec:experiments}
\begin{table*}[t!]
\centering
\small
\setlength{\tabcolsep}{0.8mm}
\begin{tabular}{lcccccc}
\toprule

Model & pLDDT $\uparrow$ & Gl. RMSD (\AA) $\downarrow$ & Core RMSD (\AA) $\downarrow$ & Gl. $<$2\AA\ (\%) $\uparrow$ & Core $<$2\AA\ (\%) $\uparrow$ & F1 $\uparrow$ \\
\midrule
ProGen2 (2.7B)          & 68.49 & 7.80  & 6.05 & 43.5 & 47.4 & 53.3 \\
ESM3 (1.4B)                      & 74.51 & 10.78 & 7.65 & 23.3 & 31.8 & 40.0 \\
ESM3 (7B)                        & 77.06 & 8.93  & 5.76 & 29.6 & 45.7 & 52.5 \\
Pinal (1B)            & \textbf{86.25} & 5.33  & 5.49 & 33.5 & 52.0 & 62.2 \\
\midrule
\rowcolor{mypurple}
\model                         & 84.48 & \textbf{3.04} & \textbf{2.69} & \textbf{63.5} & \textbf{67.3} & \textbf{74.8} \\
\bottomrule
\end{tabular}
\caption{Pretrained model performance on \dataset. 
\model achieves the best performance across all the function evaluation and structure accuracy metrics, showing its strong capability of designing proteins following human instructions.}

\label{tab:stage1_main}
\end{table*}

\subsection{Experimental Setup}
\textbf{Baselines.}
We benchmark \model against three categories of baselines: natural
language-guided models, unconditional protein language models, and
task-specific design methods. Natural language-guided baselines include:
(1) \textbf{Pinal} \citep{dai2024pinal}, an encoder-decoder framework that generates
protein sequences from natural language descriptions; and (2) \textbf{ESM3} \citep{hayes2025esm3},
a multimodal generative model that reasons jointly over sequence,
structure, and function, and can be conditioned on function annotations
expressed in natural language; and (3) \textbf{InstructPro-3B}
\citep{song2025instructpro}, a natural language-guided model for
ligand-binding protein design, which we include on the ligand task as the
closest prior system in both objective and parameter count. We also compare
against (4) \textbf{ProGen2-large-BFD90}~\citep{nijkamp2022progen2}, an autoregressive protein language model
trained on evolutionary sequence data without any functional conditioning.
Task-specific baselines are: (5) \textbf{RefineGNN} \citep{jiniterative}, which co-designs antibody CDR sequence and structure through iterative graph
refinement; and (6) \textbf{Proteina-complexa} \citep{didi2026scaling}, a flow-matching generative model that jointly generates protein sequence and all-atom coordinates conditioned on a binding target. To ensure a fair comparison, we
generate protein candidates for all baseline models using their official
code. On the ligand task, all methods are scored on the subset of targets
that every method completed, so the reported numbers are directly
comparable across rows.

More implementation details are provided in Appendix~\ref{app:baselines}.

\textbf{Evaluation Metrics.}
 
We evaluate the designed protein along three dimensions: functional performance, structure reliability, and structural accuracy. 
Functional performance is
assessed using: (1) \textbf{InterProScan \citep{jones2014interproscan} F1}, which
measures whether the designed proteins belong to the desired domains. (2) \textbf{AlphaFold3 ipTM}, which measures
interface interaction geometry, with higher scores indicating more reliable
binding. (3) \textbf{DockQ} \citep{basu2016dockq}, evaluating the binding of antibody-antigen interface in the range $[0,1]$. (4) Binding
affinity predicted by \textbf{GNINA} \citep{mcnutt2021gnina} for assessing ligand-binding proteins. Structure reliability is assessed using: (5) \textbf{pLDDT} via ESMFold, which evaluates the structure stability.
Finally, structural accuracy is assessed using: (6) \textbf{Global RMSD}, the
deviation from ground truth over the full protein. (7) \textbf{Core RMSD}, which
applies an outlier-rejecting superposition to isolate whether the
structural core is recovered when peripheral segments diverge. (8) \textbf{Pocket
RMSD}, the deviation of the ligand-binding pocket. Each RMSD is reported
together with the fraction of designs below 2\AA.

Full metric definitions are provided in Appendix~\ref{app:metrics}.

\textbf{Implementation Details.}
\label{sec:implementation}
The protein decoder uses $N_p = 32$ autoregressive Transformer
\citep{vaswani2017attention} layers with 16 attention heads and embedding
dimensionality $D_p = 2{,}560$, initialized from ProGen2-large-BFD90
\citep{nijkamp2022progen2}. 
Natural-language instructions are encoded by a
three-layer causally masked text encoder and reconstructed by a three-layer
text decoder, initialized from the first and last three layers of
PubMedBERT \citep{gu2020pubmedbert}, respectively. The ligand and antigen
encoders are initialized from ChemBERTa
\citep{chithrananda2020chemberta} and \texttt{esm2\_t12\_35M\_UR50D}
\citep{lin2023esm2}, and remain frozen throughout training. The full model
contains 3.2B parameters. At inference, we use greedy
decoding for the pretraining stage and nucleus sampling with $p=0.95$ and $\tau=0.8$ for the finetuning stage. Full
architecture and training details are given in Appendix~\ref{app:model}.

\subsection{How good is the pretrained \model?}
\label{sec:exp_stage1}

\textbf{\model is able to design proteins following human instructions for arbitrary goals.}
Table~\ref{tab:stage1_main} reports pretrained model performance on \dataset. 
\model achieves the best performance across all structure accuracy evaluation metrics. 

Specifically, \model achieves the lowest RMSD and highest $<$2\AA\ success rate of any
model compared, reducing global RMSD to 3.04\AA\ from 5.33\AA\ for the
strongest baseline and reaching 63.5\% sub-2\AA\ accuracy against 43.5\%
for the next-best model. It does so while outperforming ESM3 7B, which has
more than double the parameter count.

\textbf{Generated sequences express the requested function, not only a
plausible fold.} Structural correctness alone does not confirm that a
generated protein performs the requested function. The final column of
Table~\ref{tab:stage1_main} evaluates this directly, by checking whether
the InterPro domains specified in the conditioning text are actually
present in the folded design. \model obtains the highest F1 of any model,
recovering the requested domains substantially more often than the
strongest baseline (74.8 against 62.2).

\subsection{Can \model design high-quality antibodies?}
\label{sec:exp_antibody}
We next evaluate \model's ability in designing therapeutic antibodies.
Specifically, we finetune the pretrained \model on the curated antigen-antibody complex dataset. In inference, the finetuned model \modelAB generates all
three heavy-chain CDR loops (HCDR1--3) conditioned on a given antigen and a
natural-language design blueprint.

\begin{table}[t]
\centering
\resizebox{\columnwidth}{!}{%
\begin{tabular}{lccc}
\toprule
Model & pLDDT $\uparrow$  & ipTM $\uparrow$  & DockQ $\uparrow$ \\
\midrule
\rowcolor{mygrey}
Ground Truth              & 87.92 & 0.62 & --   \\
\midrule
ESM3                       & 86.47 & 0.50 & 0.29 \\
Pinal                      & 78.65 & 0.38 & 0.17 \\
ProGen2                    & 82.35 & 0.49 & 0.28 \\
RefineGNN                  & 85.60 & 0.52 & 0.38 \\
\midrule
\rowcolor{mypurple}
\modelAB    & \textbf{86.48} & \textbf{0.55} & \textbf{0.39} \\
\bottomrule
\end{tabular}%
}
\caption{Model performance on antibody design task.}
\label{tab:stage2_antibody}
\end{table}

\begin{table*}[t!]
\centering
\small
\setlength{\tabcolsep}{1.0mm}{
\begin{tabular}{lccccc}
\toprule
Model & pLDDT $\uparrow$ & ipTM $\uparrow$ & Pocket RMSD (\AA) $\downarrow$ & Succ $<$2\AA\ (\%) $\uparrow$ & GNINA Affinity $\downarrow$ \\
\midrule
\rowcolor{mygrey}
Ground Truth        & 89.63 & 0.92 & --   & --   & -7.35 \\
\midrule
ESM3                 & 46.68 & 0.29 & 12.33 & 1.2  & -0.13  \\
Pinal                & 84.89 & 0.89 & 4.21  & 53.9 & -6.32 \\
ProGen2              & 82.76 & 0.80 & 4.14  & 66.5 & -6.96 \\
Proteina-complexa    & 73.58 & 0.62 & 11.59 & 0.1  & -5.25 \\
InstructPro-3B       & \textbf{89.09} & \textbf{0.90} & 3.40  & 67.9 & \textbf{-7.64} \\
\midrule
\rowcolor{mypurple}
\modelLIG  & 85.49 & \textbf{0.90} & \textbf{2.24} & \textbf{72.6} & -7.05 \\
\bottomrule
\end{tabular}}
\caption{Ligand-binding protein design, seen ligands. \modelLIG places the binding pocket most accurately and succeeds on the most targets. All models are evaluated on the 1{,}288 targets completed by every method.}
\label{tab:stage2_ligand_seen}
\end{table*}

Table~\ref{tab:stage2_antibody} reports the antibody design performance.
\model achieves the best performance across all metrics, even slightly better than an antibody-specific model RefineGNN.
Specifically, \modelAB obtains the highest DockQ and
ipTM than any model compared, ahead of the general protein language models by
a clear margin (0.39 DockQ against 0.29 for ESM3 and 0.28 for ProGen2) and
narrowly ahead of RefineGNN, which is built for antibody CDR design (0.39
against 0.38). \textbf{These results demonstrate \model is capable of not only following natural language instructions, but also considering biological ground information, i.e., antigen, to design a high-affinity antibody.}

\subsection{Can \model design ligand binding proteins?}
\label{sec:exp_ligand}
Finally, we evaluate \model's ligand-conditioned variant
design quality, for both ligands seen during training and, to test generalization,
a zero-shot set of unseen ligands.

\textbf{\modelLIG places the binding pocket most accurately on seen ligands.}
Table~\ref{tab:stage2_ligand_seen} reports pocket
RMSD, docking success rate, and predicted binding affinity. \modelLIG achieves
the highest success rate and the most accurate pocket placement of any model
compared, substantially outperforming ESM3 and Proteina-complexa and reducing
pocket RMSD from 3.40\AA\ to 2.24\AA\ against InstructPro-3B, the strongest
baseline. InstructPro-3B attains a lower GNINA-predicted affinity
($-7.64$ against $-7.05$), but places the pocket less accurately and succeeds
on fewer targets (67.9\% against 72.6\%).

\textbf{Pocket accuracy holds on ligands unseen during training.}
Table~\ref{tab:stage2_ligand_unseen} repeats this
evaluation on ligands held out from training. \modelLIG retains its lead on
pocket RMSD and success rate, indicating that its ligand conditioning
generalizes beyond the training distribution rather than memorizing seen
ligand chemistry. Pinal, ProGen2 and InstructPro-3B are close on this smaller
held-out set and edge \modelLIG on pLDDT, but all three place the pocket less
accurately (2.56\AA, 2.47\AA\ and 3.05\AA\ against 1.81\AA).

\begin{table*}[t]
\centering
\small
\setlength{\tabcolsep}{1.0mm}{
\begin{tabular}{lccccc}
\toprule
Model & pLDDT $\uparrow$ & ipTM $\uparrow$ & Pocket RMSD (\AA) $\downarrow$ & Succ $<$2\AA\ (\%) $\uparrow$ & GNINA Affinity $\downarrow$ \\
\midrule
\rowcolor{mygrey}
Ground Truth        & 89.01 & 0.92 & --    & --   & -4.48 \\
\midrule
ESM3                 & 51.01 & 0.36 & 10.85 & 0.0  & -0.01 \\
Pinal                & \textbf{90.51} & \textbf{0.90} & 2.56  & 78.3 & -4.47 \\
ProGen2              & 86.12 & 0.85 & 2.47  & 81.5 & -4.45 \\
Proteina-complexa    & 85.09 & 0.70 & 9.38  & 0.0  & -2.73 \\
InstructPro-3B       & 90.33 & 0.88 & 3.05  & 70.0 & \textbf{-4.81} \\
\midrule
\rowcolor{mypurple}
\modelLIG & 85.63 & 0.88 & \textbf{1.81} & \textbf{81.7} & -4.49 \\
\bottomrule
\end{tabular}}
\caption{Ligand-binding protein design for \emph{unseen} ligands, testing generalization to ligands not observed during
training. \modelLIG places the binding pocket most accurately while matching the best baseline on success rate. All models are evaluated on the 126 targets completed by every method.}
\label{tab:stage2_ligand_unseen}
\end{table*}

\section{Analysis}
\begin{figure*}[t!]
\centering
\includegraphics[width=\textwidth]{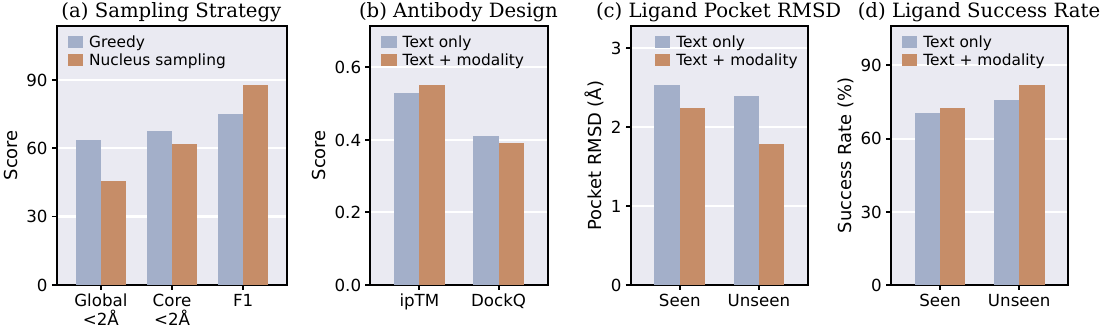}
\vspace{-0.6em}
\caption{\textbf{Analysis of decoding strategy and modality conditioning.}
(a) Greedy decoding against nucleus sampling ($p=0.95$, $\tau=0.8$) on
\dataset. (b) \modelAB\ and (c--d) \modelLIG, each with and without the
auxiliary modality encoder.}
\label{fig:analysis}
\end{figure*}

\subsection{Effect of Different Sampling Strategies}
\label{sec:analysis_sampling}

Figure~\ref{fig:analysis}(a) compares greedy decoding with nucleus sampling
($p=0.95$, $\tau=0.8$) on the pretraining test set. Greedy decoding achieves the
higher sub-2\AA\ rate, globally (63.5\% vs.\ 45.6\%) and in the core (67.3\%
vs.\ 61.6\%), while sampling achieves the higher InterProScan F1 (87.5 vs.\
74.8). The two metrics ask different questions: RMSD measures agreement with
a specific ground-truth structure, F1 only whether the requested domains are
present. Sampling produces more varied sequences, which match any single
reference less closely but express the requested function more often, and
this is not a precision--recall trade: recall increases from 64.6 to 83.5
and precision from 88.8 to 91.8.

\begin{figure*}[t!]
\centering
\includegraphics[width=\textwidth]{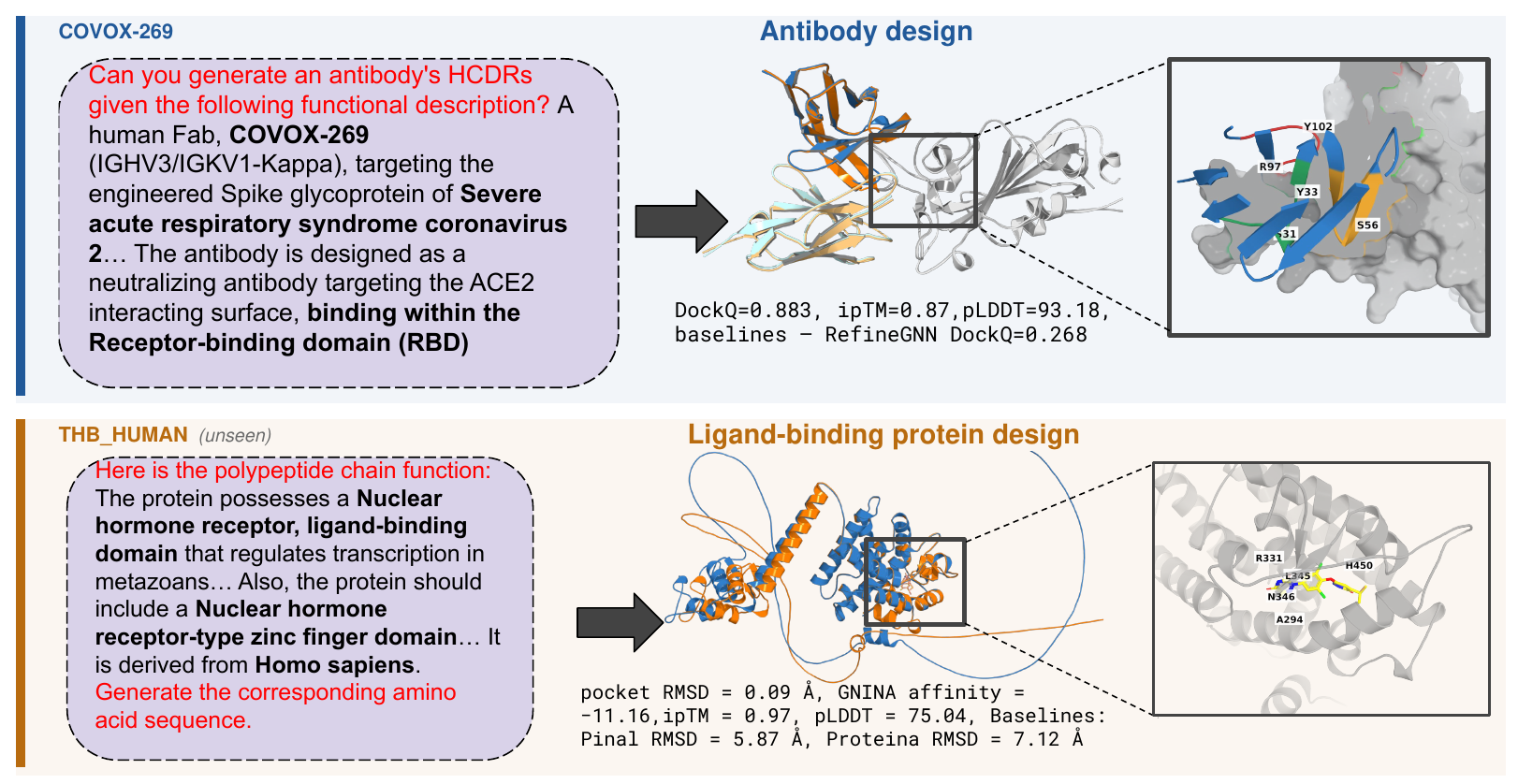}
\caption{\textbf{Case studies of the fine-tuned \model designs.} Designed structures are shown in blue, ground-truth structures in orange, and the antigen in gray. Full prompts are provided in Appendix~\ref{app:prompts}. Both cases show strong binding to the binding partner, demonstrating \model's ability to leverage both human instructions and biological context to design functional proteins.}
\label{fig:case_studies}
\end{figure*}

\subsection{Influence of Biological Grounding}

Because the blueprint already specifies the design target, we ask how much
the modality pathway adds (Figure~\ref{fig:analysis}(b--d)). On ligands the
effect is confined to pocket geometry: pocket RMSD decreases from 2.53\AA\
to 2.24\AA\ on seen ligands and from 2.41\AA\ to 1.81\AA\ on unseen ones,
while pLDDT, ipTM, and predicted affinity are largely unchanged. The
reduction is more than twice as large on unseen ligands, suggesting explicit
ligand information matters most when the target chemistry is unfamiliar,
though the unseen set is small ($N=124$). On antibodies neither model holds
a consistent advantage (DockQ 0.41 vs.\ 0.39 for text only, ipTM 0.55 vs.\
0.53 for text + modality). Both results are consistent with the blueprint
already carrying the target.

\subsection{Case Study}
\label{sec:case_studies}

Figure~\ref{fig:case_studies} shows one antibody design and one
ligand-binding protein design, illustrating how the model's conditioning is
reflected in individual designs. For COVOX-269, which targets the
SARS-CoV-2 receptor-binding domain, \modelAB recovers the interface with a
DockQ score of $0.883$, while RefineGNN ($0.268$) does
not. The instruction names the ACE2-interacting epitope and its location
within the RBD, information that is not available from the antigen
structure alone. For the unseen THB\_HUMAN target, drawn from a functional
family held out of training, \modelLIG reaches a Pocket RMSD of $0.09$~\AA{}
against $5.87$~\AA{} for Pinal and $7.12$~\AA{} for Proteina-complexa, with
a GNINA-predicted affinity of $-11.16$ that matches the $-11.12$ of the
experimental structure, indicating a pocket that is chemically as well as
geometrically accurate. These are selected examples rather than a random
sample. Four further designs are analyzed in Appendix~\ref{app:case_studies}.

\section{Conclusion}
We present \model, a functional protein design framework that integrates
natural language instructions with contextual biological knowledge through
a text encoder, dedicated ligand and antigen encoders, and a protein
decoder conditioned on both. Leveraging the composability of InterPro
annotations, we construct and release \dataset, 205.7 million $<$function
description, protein$>$ pairs spanning UniProt, enabling instruction
following across diverse protein families at a fraction of the annotation
cost. Experiments show that \model outperforms ESM3, ProGen2, and Pinal on
both structural accuracy (3.04\AA\ global RMSD vs.\ 5.33\AA) and functional
recovery (74.8 F1 vs.\ 62.2), and that the same decoder transfers to
real-world design tasks, reaching 0.39 DockQ on antibody design and a
72.6\% docking success rate on ligand-binding design. Our evaluation is
computational throughout, and experimental validation of the designed
proteins remains an important direction for future work. \model advances
natural-language-guided protein design by offering a uniform mechanism for
incorporating biological grounding, and provides practical resources for
more controllable protein generation and its real-world applications.
A potential limitation of this work is the absence of wet-lab validation, which is an essential step planned for our future research to further validate the model’s practical utility.

\section*{AI use statement}
We used generative AI in two ways in this work. First, for generating synthetic
datasets: the natural-language descriptions in \dataset and in the Stage 2
antibody corpus were produced with the aid of large language models, as described
in Section~\ref{sec:dataset} and Appendices~\ref{app:interpro}
and~\ref{app:antibody-data}. Second, for polishing writing and paraphrasing text.
Other than that, all other work was done by us. We take responsibility for the
final content of this work, including text, claims, or artifacts produced with
the aid of generative AI.

\section*{Acknowledgements}
We thank Lambda, Inc.\ for their generous compute grant
in support of this work.

\bibliography{references}
\bibliographystyle{icml2026}

\newpage
\onecolumn
\appendix
\section*{Appendix}

\newcommand{\PARAM}[1]{\textcolor{red}{\texttt{[#1]}}}

\appendix

\section{Dataset Construction}
\label{app:data}

\subsection{Stage 1: InterPro-Grounded Description Pipeline}
\label{app:interpro}

Writing a function description for each of 205.7M proteins individually is
not feasible. We instead summarize the InterPro vocabulary once and compose
per-protein descriptions from those summaries, so annotation cost scales
with the size of the vocabulary rather than the size of the corpus.

\paragraph{Entry summarization.} InterPro entries are structured
annotations rather than natural-language text. We summarize each of the
49{,}677 entries once with Gemini 3.1 Pro \citep{googledeepmind2026gemini31pro} into
two or three sentences describing what the entry is and what it does. The
summaries follow a common template so that several can be combined into one
coherent paragraph.

\paragraph{Entry selection.} A protein may carry many InterPro annotations
at differing levels of specificity. We select up to three entries per
protein across four tiers, preferring the most specific: sites, active
sites and post-translational modifications (tier 1), then domains and
repeats (tier 2), then families (tier 3), then superfamilies (tier 4).
Within a tier, entries are ordered by sequence length, which we use as a
heuristic for annotation richness, and selection across tiers is balanced so
that the retained entries describe different aspects of the protein rather
than restating one. We then remove near-duplicates by Jaccard similarity,
applied to the word sets of the summaries at a threshold of 0.3 and to the
word sets of the entry names at a threshold of 0.6. Proteins carrying no
InterPro annotation are excluded from the corpus.

\paragraph{Description assembly.} The selected summaries are joined into a
single instruction using randomized templates. Separate template sets cover
family and site descriptions, the connector phrases joining consecutive
entries, the organism, and UniProt keywords, and the assembled text is
wrapped in a randomly chosen instruction prompt. Randomizing at each
position means the model never sees a single fixed phrasing for a given
combination of annotations. Figure~\ref{fig:interpro_composability} illustrates the
pipeline for one protein, and Appendix~\ref{app:prompts} gives complete
composed descriptions.

\subsection{Span Masking and Infilling Construction}
\label{app:masking}

We construct the masked infilling task by converting 50\% of eligible training examples from standard autoregressive generation to span infilling. Sequences shorter than 10 amino acids are not used for infilling and remain autoregressive. For each eligible sequence, we first sample a target masking length uniformly between 20\% and 40\% of the sequence length.

We then divide the masked region into multiple spans. Each span length is sampled from a Poisson distribution with $\lambda=20$ and clipped to between 3 and 32 amino acids. Spans are sampled until their total length reaches the target masking length, with at most 30 spans per sequence. We require at least 5 unmasked amino acids between consecutive spans. If the sampled spans and required gaps do not fit within the sequence, spans are removed from the end until they fit. The remaining unmasked positions are then distributed randomly across the sequence, including before the first span, between spans, and after the last span.

Each masked span is replaced in the input sequence by a distinct sentinel token. We add 30 sentinel tokens, \texttt{<M1>} through \texttt{<M30>}, and a \texttt{<SEP>} token to the tokenizer vocabulary. The target consists of the masked sequence, followed by \texttt{<SEP>}, and then each sentinel token paired with the amino acid sequence that it replaced. For example, two masked spans are represented as
\texttt{[Masked Sequence] <SEP> <M1>[Span 1 Sequence]<M2>[Span 2 Sequence]}.

We also modify the natural-language instruction to match the infilling task. The functional description is extracted from the original instruction and inserted into one of 41 prompt templates that instruct the model to recover the missing sequence spans using the provided biological context. The templates vary whether this functional description appears at the beginning, middle, or end of the instruction.

Across the full training set, the resulting examples are evenly divided between autoregressive and infilling tasks, with 50.0\% assigned to each. Among infilling examples, the mean corruption ratio is 0.339 with a standard deviation of 0.075, and sequences contain an average of 6.23 masked spans. The validation and test sets show similar distributions, with mean corruption ratios of 0.337 and 0.340 and averages of 6.41 and 5.86 masked spans per sequence, respectively.

We use the same span-infilling procedure for Heavy Chain Complementarity Determining Region (HCDR) generation in Stage 2. This provides a consistent mechanism for localized sequence generation across the two training stages.

\subsection{Antibody--Antigen Dataset}
\label{app:antibody-data}
 
The antibody corpus contains 21{,}601 examples from three sources
(Table~\ref{tab:ab-splits}): a primary set of antigen-paired design blueprints,
a heavy-chain-only supplement, and an additional antigen-paired supplement.
 
\begin{table}[h]
\centering
\caption{Composition of the Stage~2 antibody corpus.}
\label{tab:ab-splits}
\small
\begin{tabular}{lrl}
\toprule
Component & Examples & Description \\
\midrule
Primary (SAbDab + blueprints) & 4{,}907  & Train 4{,}030 / Val 433 / Test 444 \\
Heavy-chain-only supplement   & 13{,}407 & Template descriptions; of 15{,}124 processed \\
Heavy+antigen supplement      & 3{,}287  & Post CDR-H3 dedup, from 5{,}316 raw \\
\midrule
\textbf{Total}                & \textbf{21{,}601} & \\
\bottomrule
\end{tabular}
\end{table}
 
\paragraph{Primary pipeline.} For each SAbDab \citep{dunbar2014sabdab} complex we
collect antigen metadata from RCSB and UniProt (name, organism, domain
annotations, functional text) together with literature context, retrieved as PMC
full text where available and falling back to the abstract or to metadata alone.
A sequence- and structure-based cleaning stage then resolves antigen chains,
rejects antibody-chain collisions using word-boundary matching with an override
list for legitimate immunoglobulin-domain antigens, detects fusion and chaperone
scaffolds (MBP, Protein A/G/L, T4 lysozyme, BRIL, SUMO, GFP, legobody and
megabody constructs) and preserves them as ordered multi-domain constructs so
the biological target is not conflated with the scaffold, and maps domains to
UniProt reference coordinates via SIFTS. Domain labels are assigned only when
overlap with an annotated UniProt Domain or Region feature reaches 80\%, and are
never inferred from model knowledge. Engineered-construct status is flagged from
purification-tag motifs alone, not from mutation count. The cleaned record is passed
to Claude Haiku 4.5 \citep{anthropic2025haiku} at temperature $0.1$ with
structured output and seven few-shot examples covering recurring edge cases.
  
\paragraph{Supplemental heavy chains.} We add heavy-chain examples without
antigen information from SAbDab complexes lacking a resolved antigen and from
TrEMBL and SwissProt \citep{uniprot2025}. Every attribute is taken from a
verified source field using audited regular expressions or lookup rules, with
each extraction tagged by the method that produced it; nothing is inferred.
Organisms map through a hand-verified adjective table, falling back to the Latin
binomial rather than inventing a demonym. Of 15{,}124 rows processed, organism
resolved for all (12{,}843 adjective, 1{,}938 binomial), germline V-gene for
11{,}640, isotype for 845, and an explicit target for only 1{,}535 --- expected,
since most heavy-chain-only entries state no target and the pipeline does not
infer one. We manually reviewed a 400-example sample across several iterations.
Attributes are assembled into descriptions by template.
 
\paragraph{CDR extraction.} Supplemental examples use SAbDab's IMGT-renumbered
structures, with CDR-H1, CDR-H2, and CDR-H3 taken from IMGT ranges 27--38,
56--65, and 105--117 \citep{lefranc2003imgt}, and ANARCI
\citep{dunbar2016anarci} used to locate the heavy-chain variable domain. We
reject entries with no identifiable heavy domain, incomplete CDRs, or a CDR-H3
shorter than five residues.
 
\paragraph{Training formulation and splits.} Antibody examples use the
span-infilling formulation of Appendix~\ref{app:masking}, with CDR regions as
infilling targets and the framework as context. We cluster CDR-H3 sequences with MMseqs2
\citep{steinegger2017mmseqs2} at 40\% sequence identity under the BLOSUM62
substitution matrix \citep{henikoff1992blosum}, then assign whole clusters to
train, validation, and test in an 8:1:1 ratio, giving 4{,}030 / 433 / 444
examples. Supplemental examples falling into a cluster already assigned to a
split are discarded under the same criterion, as are any whose PDB ID already
appears in a primary split.

\subsection{Ligand--Protein Dataset}
\label{app:ligand-data}
 
\paragraph{Data source and splits} The ligand-conditioned dataset is built on
InstructProBench \citep{song2025instructpro}. We inherit both the
protein--ligand pairs and the split assignment from that benchmark. This gives 8{,}727{,}514 training, 3{,}263 validation,
1{,}500 seen-ligand test, and 127 unseen-ligand test examples.
 
\paragraph{Function descriptions.} InstructProBench pairs each protein with text
derived from its UniProt function field alone. We replace this with a
description from the InterPro-grounded pipeline of Appendix~\ref{app:interpro},
so that Stage~2 conditions on the same description distribution seen throughout
Stage~1. This holds the text pathway fixed across stages and leaves the ligand
cross-attention pathway as the only new conditioning signal introduced in
Stage~2.

\paragraph{Evaluated subsets.} Reported results cover fewer examples than the
test splits contain. We evaluate the subset on which every compared model
produced a scoreable structure, so that all rows of a table are computed over an
identical set of targets; the common-set size is therefore limited by the
baseline with the fewest completed generations. This gives 126 of 127
unseen-ligand examples and 1{,}288 of 1{,}500 seen-ligand examples.

\section{Model and Training Details}
\label{app:model}

\subsection{Parameter Accounting and Training Configuration}
\label{app:params}

Table~\ref{tab:param-accounting} gives a component-level breakdown of
\model's parameters. The full model contains 3.20B parameters. Stage~1 trains
the protein decoder, the text encoder, the text projection, and the text
reconstruction head jointly; the adapters and modality projections do not yet
exist at this stage, so the Stage~1 model contains 2.99B parameters of which
2.87B are updated. Stage~2 initializes from the Stage~1 checkpoint, introduces
the gated cross-attention adapters and modality projections, and updates every
component except the text encoder and the two pretrained modality encoders,
giving 3.04B trainable parameters (94.9\% of the total).
 
We hold the text encoder, ChemBERTa, and ESM-2 frozen during Stage~2 so that the
conditioning representations remain fixed throughout finetuning: the model
learns to read a stable textual, chemical, or antigen embedding space rather
than co-adapting the encoder to the generation objective, which both preserves
the domain knowledge in those pretrained encoders and keeps the interface
uniform across modalities.

\begin{table}[h]
\centering
\caption{Component-level parameter counts, measured from the released Stage~2
checkpoint. ``S1'' and ``S2'' indicate whether a component is updated in
Stage~1 (pretraining) and Stage~2 (finetuning). The adapters and modality
projections are introduced in Stage~2 and therefore do not exist during
Stage~1; the Stage~1 model accordingly contains 2{,}990{,}576{,}029 parameters,
of which 2{,}873{,}132{,}124 are trainable.}
\label{tab:param-accounting}
\small
\begin{tabular}{lrcc}
\toprule
Component & Parameters & S1 & S2 \\
\midrule
ProGen2-large-BFD90 backbone (32 layers, $d{=}2560$) & 2{,}648{,}232{,}992 & \checkmark & \checkmark \\
Protein LM head                                      & 131{,}123{,}200     & \checkmark & \checkmark \\
Text encoder (PubMedBERT, first 3 layers)            & 45{,}102{,}336      & \checkmark & frozen \\
Text projection $W_p$ ($768 \rightarrow 2560$)       & 1{,}968{,}640       & \checkmark & \checkmark \\
Text reconstruction head (3 layers + LM head)        & 46{,}704{,}956      & \checkmark & \checkmark \\
\midrule
Gated cross-attention adapters ($8 \times$)          & 209{,}838{,}088     & --         & \checkmark \\
Modality projections, type embeddings, scales        & 3{,}205{,}122       & --         & \checkmark \\
\midrule
Ligand encoder (ChemBERTa, 6 layers)                 & 83{,}450{,}880      & frozen     & frozen \\
Antigen encoder (ESM-2 35M, 12 layers, $d{=}480$)    & 33{,}993{,}025      & frozen     & frozen \\
\midrule
\textbf{Total}                                       & \textbf{3{,}203{,}619{,}239} & & \\
\textbf{Trainable (Stage 2)}                         & \textbf{3{,}041{,}072{,}998} & & \\
\bottomrule
\end{tabular}
\end{table}

\paragraph{Adapter placement.} Gated cross-attention adapters are inserted after
transformer blocks 3, 7, 11, 15, 19, 23, 27, and 31, i.e.\ every fourth block,
for eight adapters in total. Each adapter applies layer normalization followed by
multi-head cross-attention into the multimodal context $C$, with its output
scaled by $\tanh(\alpha_\ell)$ for a learned scalar $\alpha_\ell$ initialized to
zero, so that the modality pathway contributes no signal at the start of Stage~2
and the Stage~1 computation is exactly preserved.

\paragraph{Modality encoders.} Ligand SMILES are encoded with ChemBERTa
\citep{chithrananda2020chemberta} (6-layer RoBERTa, $d{=}768$). Antigen sequences
are encoded with \texttt{esm2\_t12\_35M\_UR50D} \citep{lin2023esm2} (12 layers,
$d{=}480$). Each encoder output is projected to the backbone width $2560$,
rescaled by a learned per-modality scalar $s_m$, and offset by a learned
modality-type embedding $e_m$, as in Eq.~\ref{eq:eq3}. All parameters are
maintained in \texttt{float32}.

\subsection{Optimization and Compute}
\label{app:compute}

\paragraph{Stage 1 pretraining.} \model is trained in \texttt{fairseq} on 120
NVIDIA B200 GPUs (15 nodes $\times$ 8 GPUs) with \texttt{c10d} distributed
data parallelism. We use Adam with $\beta = (0.9, 0.98)$, a peak learning
rate of $5 \times 10^{-4}$ under an inverse-square-root schedule with
10{,}000 warmup updates from an initial rate of $1 \times 10^{-5}$, and
gradient-norm clipping at 1.0. Training runs in bfloat16 with float32
master weights. Batches are formed by token count, capped at 8{,}000 tokens
and 32 sequences per GPU, with a gradient accumulation factor of 2, giving
an effective batch of up to 1.92M tokens per update. The maximum sequence length is 1{,}024 positions, sequences exceeding this
length are excluded from training, validation, and test rather than
truncated, so all splits are drawn from the same length distribution. The combined objective weights the protein
term at 1.0 and the text reconstruction term at 0.5
(Section~\ref{sec:architecture}).

\paragraph{Stage 2 finetuning.} Both Stage 2 variants initialize from the
Stage 1 checkpoint and keep the Adam settings above. We reduce the peak
learning rate to $5 \times 10^{-5}$ under the same inverse-square-root
schedule, with 2{,}000 warmup updates from an initial rate of
$1 \times 10^{-6}$, and tighten gradient-norm clipping to 0.5. Batches are
capped at 1{,}100 tokens and 20 sequences per GPU with no gradient
accumulation. Adapter and modality-projection parameters are trained at
twice the base learning rate.

\section{Evaluation Details}
\label{app:eval}

\subsection{Metric Definitions}
\label{app:metrics}

All metrics are computed on predicted structures. Stage~1 monomers are
folded with ESMFold \citep{lin2023esm2} and Stage~2 complexes with
AlphaFold3 \citep{abramson2024accurate}. 

\subsubsection{Stage 1: Monomer Design}

\paragraph{Confidence.} pLDDT is the mean over C$\alpha$ B-factors of the
design's predicted structure, where ESMFold stores its per-residue
confidence.

\paragraph{Global RMSD.} Computed with PyMOL's \texttt{super}, a
sequence-independent structural superposition with outlier rejection
disabled, so every matched atom pair contributes. We use a
sequence-independent method because a design is not required to match its
target in sequence, only in structure.

\paragraph{Core RMSD.} Computed with PyMOL's \texttt{align}, which builds a
sequence alignment and fits the corresponding atoms, using five refinement
cycles and a 2.0\,\AA{} rejection cutoff. Each cycle discards the
worst-deviating atom pairs and refits, so the value describes the structural
core rather than the whole chain. Both Global and Core RMSD are computed over
all atoms of the matched residues.

\paragraph{Reference selection.} Validation and test clusters are
represented by a single sequence, selected by reservoir sampling, which
supplies the functional description. Each design is then compared against
every ground-truth structure available in its cluster, up to five, and we
report the minimum RMSD. The minimum is taken independently for the two
measures.

\paragraph{Success rates.} Gl.~$<$2\AA\ and Core~$<$2\AA\ are the fractions
of designs whose corresponding RMSD falls below 2.0\,\AA.

\paragraph{Functional recovery.} We run InterProScan
\citep{jones2014interproscan} version 5.77-108.0 on each designed sequence and
compare the entries it returns against those used to compose the prompt.
Precision is the fraction of returned entries that were requested, recall
is the fraction of requested entries that were returned, and F1 is their
harmonic mean.

\subsubsection{Stage 2: Complex Design}

\paragraph{Confidence.} ipTM is taken from AlphaFold3's summary output for
the predicted complex. pLDDT is averaged over the designed chain only,
excluding the antigen, light chain, or ligand supplied as context.

\paragraph{Pocket RMSD.} We define each structure's pocket as the protein
residues within 8\,\AA\ of its own ligand, superimpose the design onto the
reference using those residues, and report the RMSD between the two ligand
poses. Ligands are compared over heavy atoms, matched by element group so
the result does not depend on atom ordering. Pocket RMSD therefore measures
where the ligand sits rather than how the protein folds.

\paragraph{Docking success.} Succ~$<$2\AA\ is the fraction of designs with a
pocket RMSD below 2\,\AA.

\paragraph{Predicted affinity.} We rescore each predicted complex with GNINA
\citep{mcnutt2021gnina} using \texttt{--score\_only}, which scores the pose
as predicted rather than redocking. We report GNINA's empirical affinity in
kcal/mol, where lower values indicate stronger predicted binding.

\paragraph{DockQ.} Antibody designs are scored with DockQ v2.1.3
\citep{basu2016dockq}, which evaluates one chain pair at a time. We compute
the pairwise interfaces among heavy chain, light chain, and antigen, and
report the average over all chain-pair interfaces in the
complex.

\subsection{Baseline Configurations}
\label{app:baselines}

\paragraph{ESM3.} We extract the InterPro entries used to compose each prompt,
together with their residue ranges, and supply them to ESM3 as function
annotations alongside a fully masked sequence of the ground-truth length.
Conditioned on these annotations, we generate an initial sequence, then predict
a structure from it, then clear the sequence track and regenerate the sequence
from the predicted structure and the annotations. We use the SDK's default
sampling settings and evaluate both ESM3-open (1.4B) and \texttt{esm3-medium}
(7B) through the Forge API. ESM3 conditions on structured annotations rather
than free text, so unlike the other baselines it receives the target length and
residue-level positions for each annotation. We report it under these conditions
because withholding them would not represent the model fairly.

\paragraph{ESM3 annotation coverage.} ESM3's function track recognizes only a
subset of InterPro, so it rejects some of the entries our prompts specify. When
an entry is rejected we drop it and resubmit, repeating until the request is
accepted or no entries remain. Which entries survive therefore varies from
example to example, and averaging F1 over all generations would mix designs
conditioned on the full specification with designs conditioned on part of it or
none. We report ESM3's F1 in Table~\ref{tab:stage1_main} over the subset that
received every requested entry ($N=278$ for the 1.4B model and $406$ for the
7B model), which is the setting most favorable to ESM3. This restriction
applies only to ESM3's F1; ProGen2 receives no functional specification by
design and is reported over all generations as an unconditioned reference, as
are all other cells in the table.

\paragraph{Pinal.} We use the released T2struc-1.2B checkpoint and the authors'
\texttt{PinalDesign} routine, generating one sequence per prompt at temperature
$0.1$ with \texttt{mut} SaProt sampling. Pinal takes the natural-language
description directly and receives no length or positional information.

\paragraph{ProGen2.} ProGen2 has no mechanism for conditioning on function, so
we prompt it with the first 32 residues of the target sequence (fewer for
proteins shorter than 32 residues) and decode greedily until the generated
sequence matches the target length. We use the \texttt{progen2-BFD90}
checkpoint. ProGen2 therefore receives no functional specification at all, and
instead a prefix of the answer; it serves as a reference for what an
unconditioned protein language model recovers from evolutionary statistics
alone.

\paragraph{RefineGNN.} We ran inference only from the authors' released checkpoints with no additional
training or fine-tuning. The three heavy-chain CDRs are generated in three
independent passes, each loading its corresponding released model
(\texttt{RefineGNN-hcdr1}, \texttt{-hcdr2}, \texttt{-hcdr3}). Each pass redesigns a
single CDR conditioned on the remaining antibody framework and its 3D structure.
Architecture follows the released configuration (hidden dimension $256$, $4$
message-passing layers, $k = 9$ nearest neighbors, framework block size $8$).

\paragraph{Proteina-Complexa.} We use the released \texttt{complexa\_ligand.ckpt}
checkpoint and run the ligand-binder generation pipeline once per target with its
default configuration. Guidance was disabled and no reward model was active, so
generation is unguided and the method's inference-time search is not exercised;
we report it here as a generative baseline only. Each target ligand is supplied as
a protonated (pH~7.4), embedded 3D conformer (RDKit ETKDGv3, MMFF-optimized), and we evaluate one
design per target.

\paragraph{Evaluation set.} Generation fails for a small number of targets under
each baseline. We evaluate every model on the intersection of targets for which
all models produced a sequence. Where
several evaluation records share a target protein, we generate one baseline
sequence and reuse it across them.

\subsection{Additional Case Studies}
\label{app:case_studies}

\begin{figure}[t!]
\centering
\includegraphics[width=13cm]{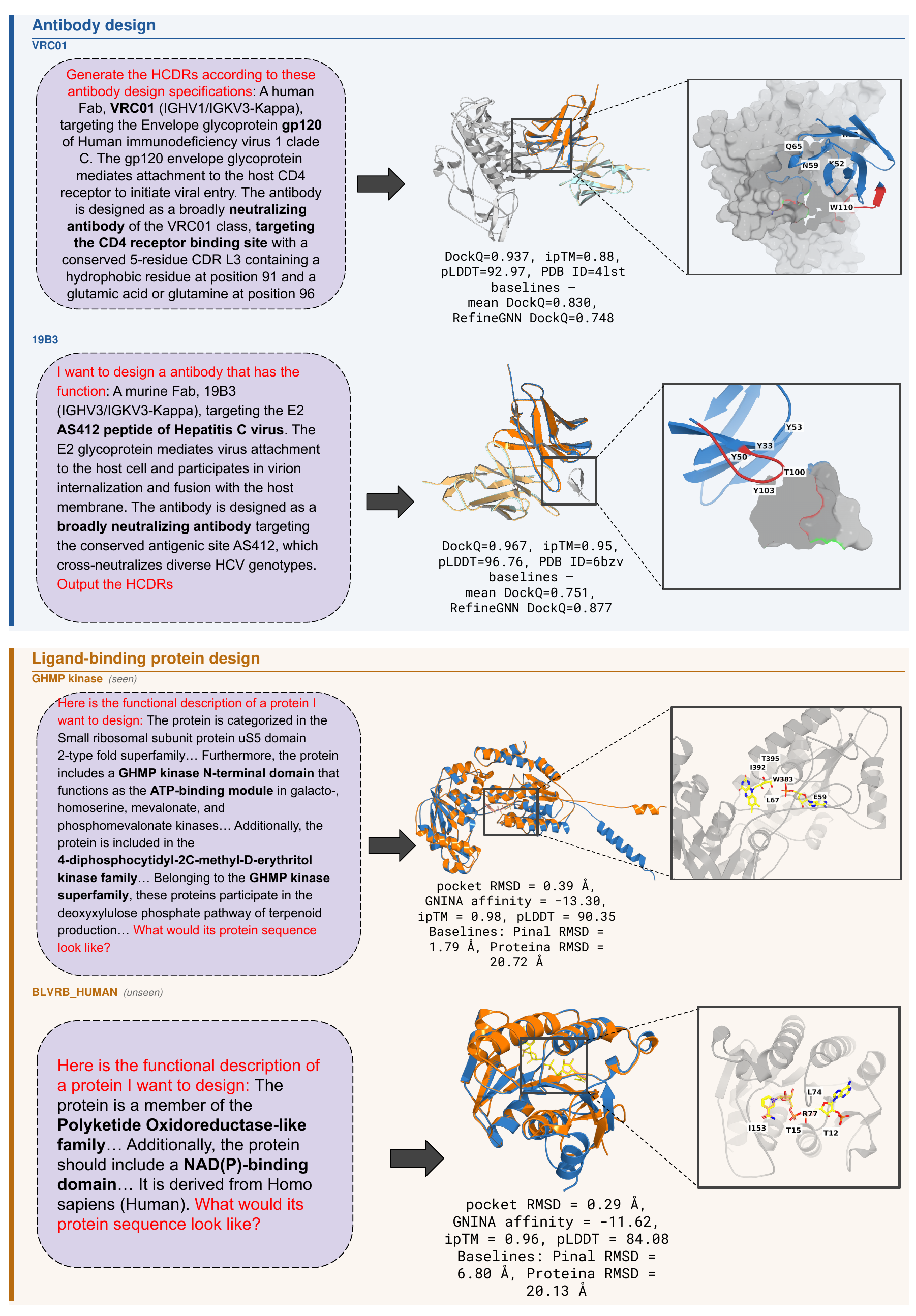}
\caption{\textbf{Additional case studies of \modelAB or \modelLIG designs.}
Each row shows the natural-language input prompt (left), the designed
structure superimposed on the ground-truth structure (middle), and a
close-up of the binding region (right). Designed structures are shown in blue and ground-truth
structures in orange; the antigen is shown in grey in the antibody
designs. Ligand-binding designs are marked as drawn from the seen or
unseen evaluation split. Ligand prompts are abbreviated (\ldots) for space,
with complete descriptions in Appendix~\ref{app:prompts}.}
\label{fig:case_studies_appendix}
\end{figure}

Figure~\ref{fig:case_studies_appendix} shows four further designs beyond
those in Section~\ref{sec:case_studies}: two antibody designs and two
ligand-binding protein designs. These examples span a wider range of
baseline performance than the two in the main text.

\textbf{Antibody design.}
VRC01, a broadly neutralizing anti-gp120 antibody, is among the most
extensively characterized antibodies in the structural literature. \modelAB
achieves a DockQ score of $0.937$, which is ahead of RefineGNN
($0.748$), but both methods recover a broadly correct interface on
this target. For 19B3, a murine Fab targeting the conserved AS412
antigenic site of the hepatitis C virus E2 glycoprotein, \modelAB achieves
the highest DockQ score in our antibody shortlist ($0.967$), with
RefineGNN also performing well ($0.877$). Taken
together with COVOX-269 in Section~\ref{sec:case_studies}, these examples
indicate that the margin over structure-conditioned baselines varies
considerably across targets, and is largest where the instruction supplies
epitope information that is not recoverable from the antigen structure
alone.

\textbf{Ligand-binding protein design.}
On the seen GHMP kinase target, \modelLIG recovers the ATP-binding pocket
with a Pocket RMSD of $0.39$~\AA, compared with $1.79$~\AA{} for Pinal and
$20.72$~\AA{} for Proteina-complexa; its GNINA-predicted affinity is
$-13.30$ against $-13.78$ for the experimental structure. Pinal remains
within a few angstroms of the correct pocket on this target, in contrast
to the unseen targets, where both baselines fail. On the unseen
BLVRB\_HUMAN target, a human biliverdin reductase, \modelLIG achieves a
Pocket RMSD of $0.29$~\AA{} while Pinal and Proteina-complexa obtain
$6.80$~\AA{} and $20.13$~\AA, respectively, and attains a predicted
affinity of $-11.62$ against $-11.37$ for the experimental structure. We
note that the BLVRB\_HUMAN design has a pLDDT of $84.08$ against $96.90$
for the experimental structure, a larger confidence gap than in the other
ligand examples, indicating that close pocket geometry does not always
coincide with high predicted confidence across the full chain.

\definecolor{promptred}{HTML}{CC0000}
\definecolor{connectororange}{HTML}{E69138}

\newcommand{\pr}[1]{\textcolor{promptred}{#1}}
\newcommand{\cn}[1]{\textcolor{connectororange}{#1}}
\section{Full Functional Descriptions}
\label{app:prompts}

The prompts in Figure~\ref{fig:case_studies} and
Figure~\ref{fig:case_studies_appendix} are abbreviated for space. We
reproduce them in full below, exactly as presented to the model. Following
the figures, \pr{the task instruction} is shown in red and the antibody,
domain, family, and organism terms that identify the target in
\textbf{bold}. In the ligand-binding prompts, \cn{the connector phrases
that join consecutive InterPro entries} (Appendix~\ref{app:interpro}) are
additionally shown in orange. The VRC01 and 19B3 prompts appear in full in
Figure~\ref{fig:case_studies_appendix} and are not repeated here.

\paragraph{COVOX-269.}
\begin{quote}\small
\pr{Can you generate an antibody's HCDRs given the following functional
description?} A human Fab, \textbf{COVOX-269} (IGHV3/IGKV1-Kappa),
targeting the engineered Spike glycoprotein of \textbf{Severe acute
respiratory syndrome coronavirus 2}. The Spike glycoprotein attaches the
virion to the cell membrane by interacting with the host ACE2 receptor,
initiating infection and mediating fusion of viral and cellular membranes.
The antibody is designed as a \textbf{neutralizing antibody} targeting the
ACE2 interacting surface, \textbf{binding within the Receptor-binding
domain (RBD)}.
\end{quote}

\paragraph{GHMP kinase (seen).}
\begin{quote}\small
\pr{Here is the functional description of a protein I want to design:}
\cn{The protein is categorized in the} Small ribosomal subunit protein uS5
domain 2-type fold superfamily, which features a left-handed
$\beta$-$\alpha$-$\beta$ structure utilized in diverse nucleic acid-binding
proteins and GHMP family kinases. This domain serves as a key component in
translational machinery and DNA repair enzymes, such as MutL and DNA gyrase
B, while also facilitating ATP-dependent phosphoryl transfer in metabolic
enzymes like galactokinase and mevalonate kinase. \cn{Furthermore, the
protein includes a} \textbf{GHMP kinase N-terminal domain} that functions as
the \textbf{ATP-binding module} in galacto-, homoserine, mevalonate, and
phosphomevalonate kinases. It contains a conserved Gly/Ser-rich region
essential for catalyzing the phosphorylation of these substrates.
\cn{Additionally, the protein is included in the}
\textbf{4-diphosphocytidyl-2C-methyl-D-erythritol kinase family}, which
functions as a key enzyme in the non-mevalonate pathway for isoprenoid
biosynthesis. \cn{Belonging to the} \textbf{GHMP kinase superfamily}, these
proteins participate in the deoxyxylulose phosphate pathway of terpenoid
production and are essential for chloroplast development. \pr{What would its
protein sequence look like?}
\end{quote}

\paragraph{THB\_HUMAN (unseen).}
\begin{quote}\small
\pr{Here is the polypeptide chain function:} \cn{The protein possesses a}
\textbf{Nuclear hormone receptor, ligand-binding domain} that regulates
transcription in metazoans by binding specific small hydrophobic molecules
within a conserved $\alpha$-helical sandwich structure. This binding event
induces conformational changes in the ligand-dependent activation function
(AF-2) region, facilitating the recruitment of transcriptional coregulators
to modulate gene expression. \cn{Furthermore, the protein is assigned to
the} \textbf{Nuclear hormone receptor family NR1 subfamily}, which functions
as a class of ligand-activated transcription factors regulating development,
metabolism, and immune responses. Members utilize DNA-binding and
ligand-binding domains to modulate gene expression in response to endogenous
ligands such as thyroid hormones, vitamin D3, oxysterols, and bile acids.
These receptors act as transcriptional switches to control critical
physiological processes, including lipid, glucose, and cholesterol
homeostasis. \cn{Also, the protein should include a} Nuclear hormone
receptor-type zinc finger domain that functions as a DNA-binding module
composed of two C4-type zinc-nucleated motifs. These motifs determine
sequence specificity for hormone-response elements and facilitate
dimerization to regulate ligand-responsive gene transcription. \cn{It is
derived from} \textbf{Homo sapiens (Human)} \cn{and corresponds to UniProt
keyword} Zinc-finger, Transcription regulation, and Proteomics
identification. \pr{Generate the corresponding amino acid sequence.}
\end{quote}

\paragraph{BLVRB\_HUMAN (unseen).}
\begin{quote}\small
\pr{Here is the functional description of a protein I want to design:}
\cn{The protein is a member of the} \textbf{Polyketide Oxidoreductase-like
family}, which comprises oxidoreductases crucial for the biosynthesis of
secondary metabolites, including polyketides and mycotoxins such as
aflatoxins and sterigmatocystin. These enzymes catalyze specific oxidation
steps to convert intermediates into complex structures, forming functional
groups and rings essential for biological activity. Additionally, some
members demonstrate broad specificity by reducing flavins and other
substrates involved in diverse metabolic processes. \cn{Furthermore, the
protein should include a} \textbf{NAD(P)-binding domain} that adopts a core
Rossmann-type fold to bind NAD or NADP cofactors. It is a versatile
structural module found in a diverse array of oxidoreductases, such as
alcohol and lactate dehydrogenases, as well as in certain potassium channels
and transcriptional repressors. \cn{It is derived from} \textbf{Homo sapiens
(Human)} \cn{and corresponds to UniProt keyword} Cytoplasm, Proteomics
identification, and Transferase. \pr{What would its protein sequence look
like?}
\end{quote}

\end{document}